\pdfoutput=1

\documentclass[11pt,a4paper]{article}
\usepackage[hyperref]{emnlp2020}
\usepackage{times}
\usepackage{latexsym}
\usepackage{float}

\usepackage{booktabs}
\usepackage{microtype}
\usepackage{tablefootnote}
\usepackage{graphicx}
\usepackage{amssymb}
\usepackage{enumitem}
\usepackage{comment}
\usepackage{todonotes}
\usepackage{enumitem}

\aclfinalcopy 
\def\aclpaperid{1981} 

\newcommand{\colorann}[3]{\textcolor{#1}{${}^{#2}[$#3$]$}}
\definecolor{auburn}{rgb}{0.43, 0.21, 0.1}
\newcommand{\avi}[1]{\colorann{auburn}{Avi}{#1}}

 \newcommand{\AC}{Answer Corrector}

\title{Answer Span Correction in Machine Reading Comprehension}

\author{Revanth Gangi Reddy\thanks{\hspace{2mm}Work done during AI Residency at IBM Research.}, Md Arafat Sultan, Efsun Sarioglu Kayi, \\\textbf{Rong Zhang, Vittorio Castelli\thanks{\hspace{2mm}Corresponding author.}, Avirup Sil} \\
IBM Research AI \\
  \texttt{g.revanthreddy111@gmail.com}, \texttt{arafat.sultan@ibm.com},\\ \texttt{\{avi,zhangr,vittorio\}@us.ibm.com}, \texttt{efsun@gwu.edu} \\}

\date{}

\begin{document}
\maketitle
\begin{abstract}

Answer validation in machine reading comprehension (MRC) consists of verifying an extracted answer against an input context and question pair.
Previous work has looked at re-assessing the ``answerability'' of the question given the extracted answer.
Here we address a different problem: the tendency of existing MRC systems to produce \emph{partially correct} answers when presented with answerable questions.
We explore the nature of such errors and propose a post-processing correction method that yields  statistically significant performance improvements over state-of-the-art MRC systems in both monolingual and multilingual evaluation.  

\end{abstract}

\section{Introduction }
\begin{figure}
\centering
\includegraphics[width=7.7cm]{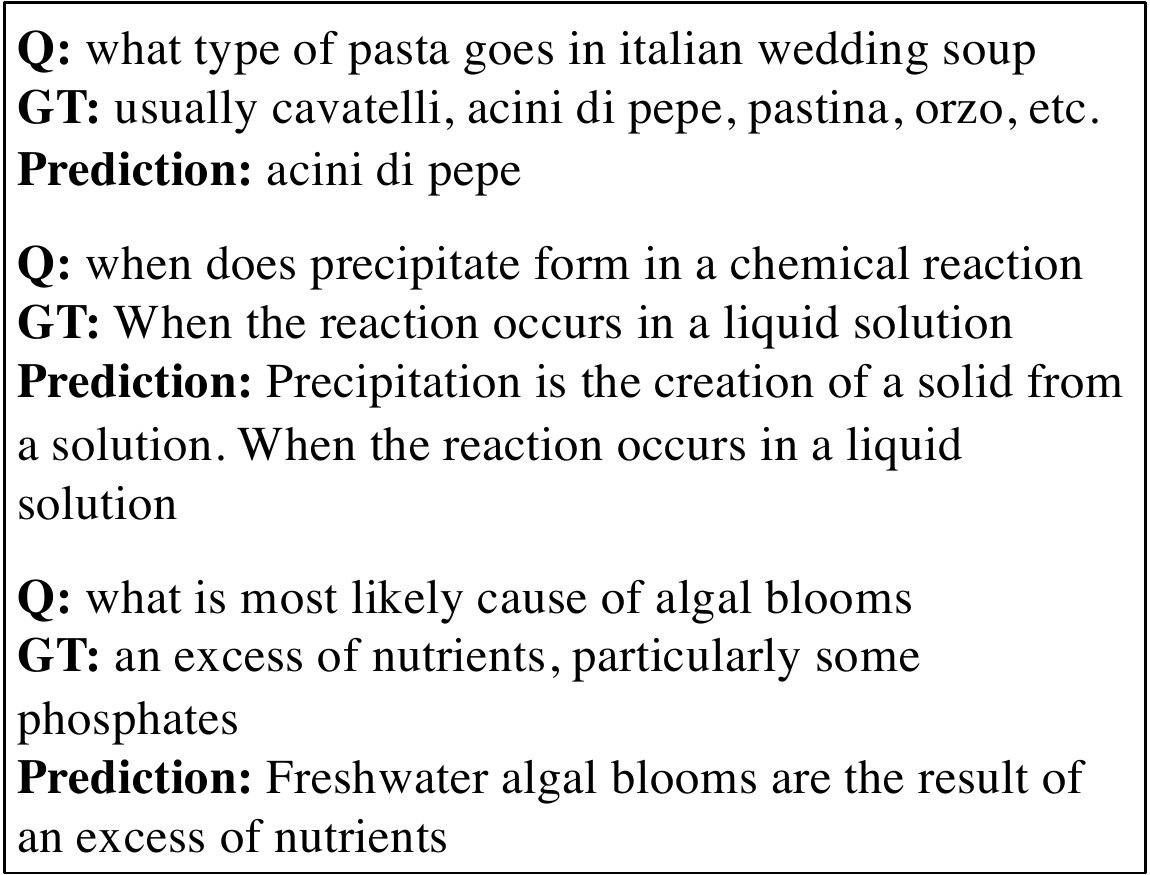}
\caption{Examples of partially correct MRC predictions and corresponding ground truth (GT) answers. 
The reader fails to find a minimal yet sufficient answer in all three cases.}
\label{figure:error-examples}
\end{figure}

Extractive machine reading comprehension (MRC) has seen unprecedented progress in recent years \cite{pan2019frustratingly,liu2020rikinet,khashabi2020unifiedqa, lewis2019mlqa}.
 Nevertheless, existing MRC systems---\textbf{\textit{readers}}, henceforth---extract only partially correct answers in many cases.
At the time of this writing, for example, the top systems on leaderboards like SQuAD~\cite{ rajpurkar-etal-2016-squad}, HotpotQA~\cite{yang2018hotpotqa} and Quoref~\cite{dasigi2019quoref} all have a difference of 5--13 points between their exact match (EM) and F1 scores, which are measures of full and partial overlap with the ground truth answer(s), respectively.
Figure~\ref{figure:error-examples} shows three examples of such errors that we observed in a state-of-the-art (SOTA) RoBERTa-large \cite{liu2019roberta} model on the recently released Natural Questions (NQ) \cite{kwiatkowski2019natural} dataset. 
In this paper, we investigate the nature of such partial match errors in MRC and also their post hoc correction in context.  

Recent work on answer validation \cite{penas2007overview} has focused on improving the prediction of the answerability of a question given an already extracted answer. 
\citeauthor{Hu2019ReadV}~\shortcite{Hu2019ReadV} look for support of the extracted answer in local entailments between the answer sentence and the question.
\citeauthor{back2020neurquri}~\shortcite{back2020neurquri} propose an attention-based model that explicitly checks if the candidate answer satisfies all the conditions in the question.
\citeauthor{zhang2020retrospective}~\shortcite{zhang2020retrospective} use a two-stage reading process: a sketchy reader produces a preliminary judgment on answerability and an intensive reader extracts candidate answer spans to verify the answerability.  

Here we address the related problem of \emph{improving the answer span}, and  present a correction model that re-examines the extracted answer in context to suggest corrections.
Specifically, we mark the extracted answer with special delimiter tokens and show that a corrector with architecture similar to that of the original reader can be trained to produce a new accurate prediction.

Our main contributions are as follows: 
\textbf{(1)} We analyze partially correct predictions of a SOTA English reader model, revealing a distribution over three broad categories of errors. 
\textbf{(2)} We show that an \AC\ model can be trained to correct errors in all three categories given the question and the original prediction in context. 
\textbf{(3)} We further show that our approach generalizes to other languages: our proposed answer corrector
yields statistically significant improvements over strong RoBERTa and Multilingual BERT (mBERT) \cite{devlin-etal-2019-bert} baselines on both monolingual and multilingual benchmarks.

\section{Partial Match in MRC}
\label{section:partial-match}
Short-answer extractive MRC only extracts short sub-sentence answer spans, but locating the best span can still be hard. 
For example, the answer may contain complex substructures including multi-item lists or question-specific qualifications and contextualizations of the main answer entity.
This section analyzes the distribution of broad categories of errors that neural readers make when they fail to pinpoint the exact ground truth span (GT) despite making a partially correct prediction.

\begin{table}[t]
\centering
\begin{tabular}{lr}
\hline
\multicolumn{1}{c}{Error} & \multicolumn{1}{c}{\%} \\
\hline
Single-Span GT & 67\\
~~\textit{Prediction $\subset$ GT} & \textit{33} \\
~~\textit{GT $\subset$ Prediction} & \textit{28} \\
~~\textit{Prediction $\cap$ GT} $\neq \emptyset$ & \textit{6} \\
\hline
Multi-Span GT & 33 \\
\hline
\end{tabular}
\caption{Types of errors in NQ dev predictions with a partial match with the ground truth.
} 
\label{tab:ErrorCategories}
\end{table}


To investigate, we evaluate a RoBERTa-large reader (details in Section~\ref{section:method}) on the NQ dev set and identify 587 examples where the predicted span has only a partial match (EM = 0, F1 $>$ 0) with the GT. Since most existing MRC readers are trained to produce single spans, we discard examples where the NQ annotators provided multi-span answers consisting of multiple non-contiguous subsequences of the context. After discarding such multi-span GT examples, we retain 67\% of the 587 originally identified samples.

There are three broad categories of partial match errors:
\begin{enumerate}[noitemsep, leftmargin=*]
\item\textbf{Prediction $\subset$ GT}: As the top example in Figure~\ref{figure:error-examples} shows, in these cases, the reader only extracts part of the GT and drops words/phrases such as items in a comma-separated list and qualifications or syntactic completions of the main answer entity.
\item\noindent\textbf{GT $\subset$ Prediction}: Exemplified by the second example in Figure~\ref{figure:error-examples}, this category comprises cases where the model's prediction subsumes the closest GT, and is therefore not minimal.
In many cases, these predictions lack syntactic structure and semantic coherence as a textual unit.
\item\textbf{Prediction $\cap$ GT $\neq \emptyset$}: This final category consists of cases similar to the last example of Figure~\ref{figure:error-examples}, where the prediction partially overlaps with the GT. (We slightly abuse the set notation for conciseness.) Such predictions generally exhibit both verbosity and inadequacy.
\end{enumerate}

Table~\ref{tab:ErrorCategories} shows the distribution of errors over all categories.



\section{Method}
In this section, we describe our approach to correcting partial-match predictions of the reader.
\label{section:method}

\subsection{The Reader}
We train a baseline reader for the standard MRC task of answer extraction from a passage given a question.
The reader uses two classification heads on top of a pre-trained transformer-based language model~\cite{liu2019roberta}, pointing to the start and end positions of the answer span.
The entire network is then fine-tuned on the target MRC training data.
For additional details on a transformer-based reader, see \cite{devlin-etal-2019-bert}.

\begin{figure}[t!]
\centering
  \includegraphics[width=7.65cm]{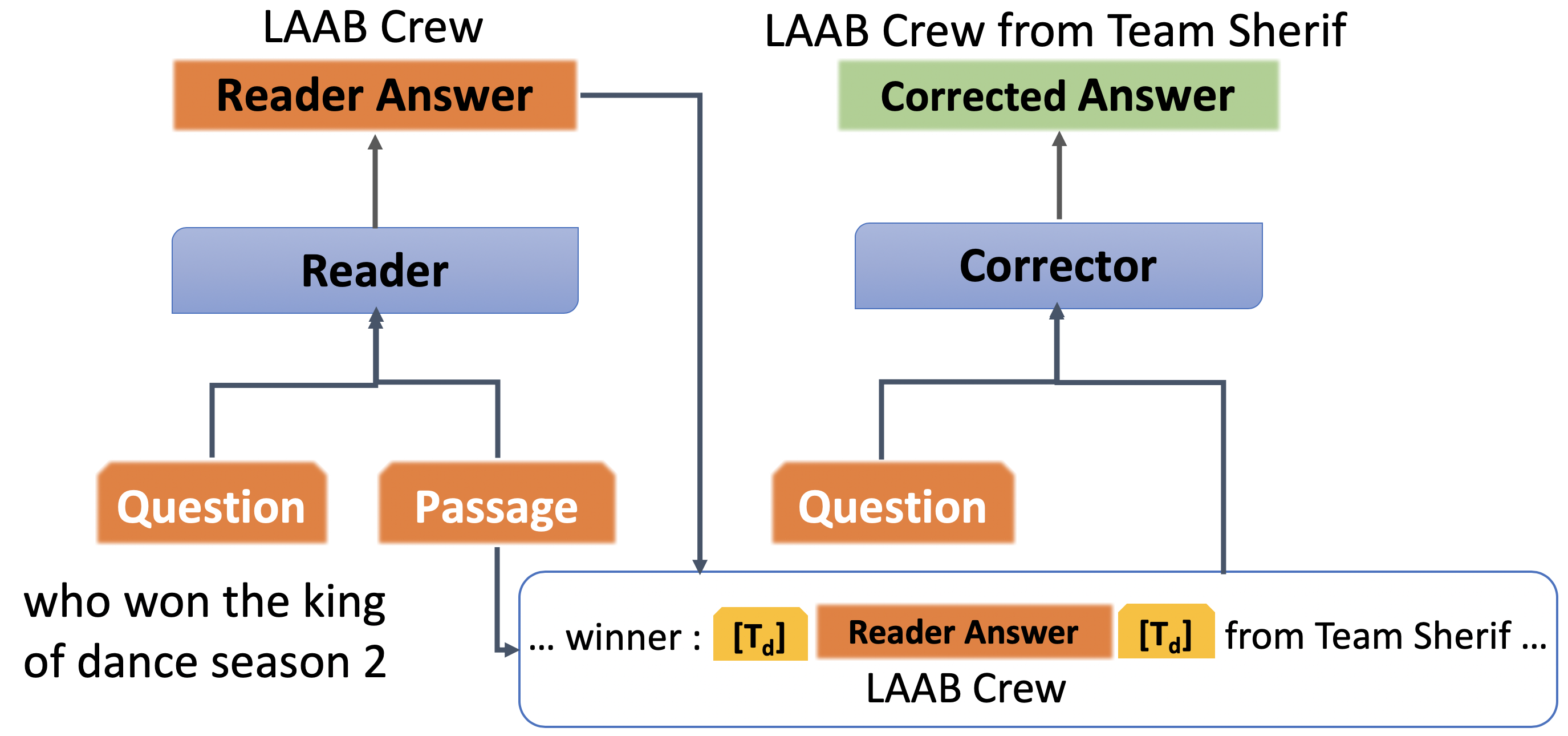}
  \caption{Flow of an MRC instance through the reader-corrector pipeline. The corrector takes an input, with special delimiter tokens ([$T_d$]) marking the  reader's predicted answer in context, and makes a new prediction.}
  \label{fig:tbp}
\end{figure}

\subsection{The Corrector}
Our correction model uses an architecture that is similar to the reader's, but takes a slightly different input.
As shown in Figure~\ref{fig:tbp}, the input to the corrector contains special delimiter tokens marking the boundaries of the reader's prediction, while the rest is the same as the reader's input. 
Ideally, we want the model to keep answers that already match the GT intact and correct the rest.

To generate training data for the corrector, we need a reader's predictions for the training set. To obtain these, we split the training set into five folds, train a reader on four of the folds and get predictions on the remaining fold.
We repeat this process five times to produce predictions for all (question, answer) pairs in the training set. The training examples for the corrector are generated using these reader predictions and the original GT annotations.
To create examples that do not require correction, we create a new example from each original example where we delimit the GT answer itself in the input, indicating no need for correction.
For examples that need correction, we use the reader's top $k$ incorrect predictions ($k$ is a hyperparameter) to create an example for each, where the input is the reader's predicted span and the target is the GT.
The presence of both GT (correct) and incorrect predictions in the input data ensures that the corrector learns both to detect errors in the reader's predictions and to correct them.




\section{Experiments}

\subsection{Datasets}
\label{datasets}
We evaluate our answer correction model on two benchmark datasets. \newline

\noindent \textbf{Natural~Questions} (NQ) \cite{kwiatkowski2019natural} is an English MRC benchmark which contains questions from Google users, and requires systems to read and comprehend entire Wikipedia articles. We evaluate our system only on the answerable questions in the dev and test sets. NQ contains 307,373 instances in the train set, 3,456 answerable questions in the dev set and 7,842 total questions in the blind test set of which an undisclosed number is answerable.
To compute exact match on answerable test set questions, we submitted a system that always outputs an answer and took the recall value from the leaderboard.\footnote{\href{https://ai.google.com/research/NaturalQuestions/leaderboard}{\textit{ai.google.com/research/NaturalQuestions/leaderboard}}}\newline

\begin{table}[t]
\centering
\begin{tabular}{lcc}
\hline
\multicolumn{1}{c}{Model} & Dev & Test \\
\hline
RoBERTa Reader & 61.2 & 62.4\\
+ Corrector & \textbf{62.8} & \textbf{63.7}  \\
\hline
\end{tabular}
\caption{Exact Match results on Natural Questions. } 
\label{tab:NQTable}
\end{table}

\noindent \textbf{MLQA} 
\cite{lewis2019mlqa} is a multilingual extractive MRC dataset with monolingual and cross-lingual instances in seven languages: English (en), Arabic (ar), German (de) , Spanish (es), Hindi (hi), Vietnamese (vi) and Simplified Chinese (zh). It has 15,747 answerable questions in the dev set and a much larger test set with 158,083 answerable questions.

\subsection{Setup}

Our NQ and MLQA readers fine-tune a RoBERTa-large and an mBERT (cased, 104 languages) language model, respectively.
Following  \citeauthor{alberti2019bert}~\shortcite{alberti2019bert}, we fine-tune the RoBERTa model first on SQuAD2.0 \cite{rajpurkar-etal-2018-know} and then on NQ. Our experiments showed that training on both answerable and unanswerable questions yields a stronger and more robust reader for NQ, even though we evaluate only on answerable questions. For MLQA, we follow \citeauthor{lewis2019mlqa}~\shortcite{lewis2019mlqa} to train on SQuAD1.1  \cite{rajpurkar-etal-2016-squad}, as MLQA does not contain any training data. We obtain similar baseline results as reported in \cite{lewis2019mlqa}. All our implementations are based on the Transformers library by \citeauthor{Wolf2019HuggingFacesTS}~\shortcite{Wolf2019HuggingFacesTS}.

For each dataset, the answer corrector uses the same underlying transformer language model as the corresponding reader. 
While creating training data for the corrector, to generate examples that need correction, we take the two ($k=2$) highest-scoring incorrect reader predictions (the value of $k$ was tuned on dev).
Since our goal is to fully correct any inaccuracies in the reader's prediction, we use exact match (EM) as our evaluation metric. 
We train the corrector model for one epoch with a batch size of 32, a warmup rate of 0.1 and a maximum query length of 30. For NQ, we use a learning rate of 2e-5 and a maximum sequence length of 512; the corresponding values for MLQA are 3e-5 and 384, respectively.

\subsection{Results}
We report results obtained by averaging over three seeds. 
Table \ref{tab:NQTable} shows the results on the answerable questions of NQ.
Our answer corrector 
improves upon the reader by 1.6 points on the dev set and 1.3 points on the blind test set.

\begin{table}[h]
\centering
\begin{tabular}{l|cc|cc}
\multicolumn{1}{c}{} & \multicolumn{2}{c}{En-Context} & \multicolumn{2}{c}{G-XLT} \\
\hline
\multicolumn{1}{c|}{Model} & Dev & Test & Dev & Test                          \\ 
                                                           
\hline
mBERT Reader & 47.5 & 45.6 & 35.0 & 34.7  \\
+ Corrector  & \textbf{48.3} & \textbf{46.4} & \textbf{35.5} & \textbf{35.3}  \\
\hline
\end{tabular}
\caption{Exact match results on MLQA. En-Context refers to examples with an English paragraph, G-XLT refers to the generalized cross-lingual transfer task. } 
\label{tab:MLQATable}
\end{table}

Results on MLQA are shown in Table~\ref{tab:MLQATable}.
We compare performances in two settings: one with the paragraph in English and the question in any of the seven languages (En-Context), and the other being the Generalized Cross-Lingual task (G-XLT) proposed in \cite{lewis2019mlqa}, where the final performance is the average over all 49  (question, paragraph) language pairs involving the seven languages.

\begin{table}[h]
\small
\center
\resizebox{\columnwidth}{!}{\begin{tabular}{l|ccccccc}
    \hline
    q\textbackslash c &	en&	es&	hi&	vi&	de&	ar&	zh\\
\midrule
    en &	0.2$\uparrow$	&$\downarrow$0.1&	$\downarrow$0.2&	$\downarrow$0.4&	$\downarrow$0.1&	$\downarrow$0.1& $\downarrow$0.3 \\
    es&	0.9$\uparrow$&	$\downarrow$0.2&	$\downarrow$0.1	&0.2$\uparrow$&	0.8$\uparrow$	&0.5$\uparrow$&	1.4$\uparrow$ \\
    hi&	0.8$\uparrow$&	0.8$\uparrow$&	0.8$\uparrow$&	0.8$\uparrow$&	0.6$\uparrow$&	0.4$\uparrow$& 0.2$\uparrow$ \\
    vi&	0.9$\uparrow$&	1.7$\uparrow$&	0.7$\uparrow$&	0.3$\uparrow$&	1.3$\uparrow$&	0.9$\uparrow$&	0.5$\uparrow$ \\
    de&	1.7$\uparrow$&	0.6$\uparrow$&	$\downarrow$0.1&	0.6$\uparrow$&	0.1$\uparrow$&	1.3$\uparrow$&	0.9$\uparrow$ \\
    ar&	0.5$\uparrow$&	1.0$\uparrow$&	0.4$\uparrow$&	0.7$\uparrow$&	0.9$\uparrow$&	0.5$\uparrow$&	0.4$\uparrow$ \\
    zh&	0.9$\uparrow$&	0.1$\uparrow$&	0.9$\uparrow$&	0.8$\uparrow$&	1.3$\uparrow$&	0.4$\uparrow$&	0.3$\uparrow$ \\
\midrule
    \textbf{AVG}&	\textbf{0.8$\uparrow$}&	\textbf{0.6$\uparrow$}&	\textbf{0.3$\uparrow$}&	\textbf{0.4$\uparrow$}&	\textbf{0.7$\uparrow$}&	\textbf{0.6$\uparrow$}&	\textbf{0.5$\uparrow$}\\
    \bottomrule
    \end{tabular}}
    \caption{Changes in exact match with the answer corrector, for all the language pair combinations in the MLQA test set. The final row shows the gain for each paragraph language averaged over questions in different languages.}
    \label{tab:gxlt}
\end{table}

Table \ref{tab:gxlt} shows the differences in exact match scores for all 49 MLQA language pair combinations, from using the answer corrector over the reader. On average, the corrector gives performance gains for paragraphs in all languages (last row). The highest gains are observed in English contexts, which is expected as the model was trained to correct English answers in context. However, we find that the approach generalizes well to the other languages in a zero-shot setting as exact match improves in 40 of the 49 language pairs.

We performed Fisher randomization tests \cite{fisher1936design} on the exact match numbers to verify the statistical significance of our results. 
For MLQA, we found our reader + corrector pipeline to be significantly better than the baseline reader on the 158k-example test set at $p<0.01$.
For NQ, the $p$-value for the dev set results was approximately $0.05$.

\section{Analysis}

\subsection{Comparison with Equal Parameters}
In our approach, the reader and the corrector have a common architecture, but their parameters are separate and independently learned. 
To compare with an equally sized baseline, we build an ensemble system for NQ which averages the output logits of two different RoBERTa readers. As Table \ref{tab:parameter comparison} shows, the corrector on top of a single reader still outperforms this ensemble of readers.
These results confirm that the proposed correction objective complements the reader's extraction objective well and is fundamental to our overall performance gain.


\begin{table}[t]
\centering
\begin{tabular}{cc}
\hline
Model & EM \\
\hline
Reader  & 61.2 \\
Ensemble of Readers & 62.1 \\
Reader + Corrector & \textbf{62.8}\\
\hline
\end{tabular}
\caption{Error correction versus model ensembling.} 
\label{tab:parameter comparison}
\end{table}

\subsection{Changes in Answers}
We inspect the changes made by the answer corrector to the reader's predictions on the NQ dev set.
Overall, it altered 13\% (450 out of 3,456) of the reader predictions. Of all changes, 24\% resulted in the correction of an incorrect or a partially correct answer to a GT answer and 10\% replaced the original correct answer with a new correct answer 
(due to multiple GT annotations in NQ).
In 57\% of the cases, the change did not correct the error.
On a closer look, however, we observe that the F1 score went up in more of these cases (30\%) compared to when it dropped (15\%).
Finally, 9\% of the changes introduced an error in a correct reader prediction. These statistics are shown in Table~\ref{tab:change_stats}.

\begin{table}[h]
    \vspace{1mm}
    \centering
    \begin{tabular}{|c | c  c|}
    \hline
    R\textbackslash R+C & Correct & Incorrect \\
    \hline
    Correct & 45 (10\%) & 43 (9\%)\\
    Incorrect & 109 (24\%) & 253 (57\%)\\
    \hline
    \end{tabular}
    
    \caption{Statistics for the correction model altering original reader predictions. The row header refers to predictions from the reader and the column header refers to the final output from the corrector.}
    \label{tab:change_stats}
\end{table}

Table \ref{tab:example_predictions_correction} shows some examples of correction made by the model for each of the three single-span error categories of Table \ref{tab:ErrorCategories}. Two examples wherein the corrector introduces an error into a previously correct output from the reader model are shown in Table \ref{tab:example_predictions_wrong}.

\begin{table*}[h]
    \centering
    \small
    \begin{tabular}{cp{2.4cm}p{5.0cm}p{4.2cm}}
    \hline
       \textbf{Error Class} & \textbf{Question}  & \textbf{Passage} & \textbf{Prediction} \\
       \hline
       Prediction $\subset$ GT & who won the king of dance season 2 &
       ... Title Winner : \textbf{LAAB Crew From Team Sherif} ,
           1st Runner-up  : ADS kids From Team Sherif ,
           2nd Runner-up  : Bipin and Princy From Team Jeffery ...
       & \textbf{R:} LAAB Crew
       
       ~
       
       \textbf{R+C:} LAAB Crew From Team Sherif\\
       \hline
       GT $\subset$ Prediction & unsaturated fats are comprised of lipids that contain & ... An unsaturated fat is a fat or fatty acid in which there is \textbf{at least one double bond} within the fatty acid chain. A fatty acid chain is monounsaturated if it contains one double ... &
       \textbf{R:} An unsaturated fat is a fat or fatty acid in which there is at least one double bond
       
       ~

       \textbf{R+C:} at least one double bond
       \\
       \hline
      Prediction $\cap$ GT $\neq \emptyset$ & what is most likely cause of algal blooms &... colloquially as red tides. Freshwater algal blooms are the result of \textbf{an excess of nutrients , particularly some phosphates}. The excess of nutrients may originate from fertilizers that are applied to land for agricultural or recreational ... & \textbf{R:} Freshwater algal blooms are the result of an excess of nutrients
      
      ~
      
      \textbf{R+C:} an excess of nutrients , particularly some phosphates\\
       
       \hline
    \end{tabular}
    \caption{Some examples for different error classes in the Natural Questions dev set wherein the answer corrector corrects a previously incorrect reader output. Ground truth answer is marked in bold in the passage. \textbf{R} and \textbf{C} refer to reader and corrector, respectively.}
    \label{tab:example_predictions_correction}
\end{table*}

\begin{table*}[h]
    \vspace{4mm}
    \centering
    \small
    \begin{tabular}{p{2.6cm}p{6.2cm}p{4.9cm}}
    \hline
       \textbf{Question}  & \textbf{Passage} & \textbf{Prediction} \\
       \hline
       	where are the cones in the eye located
       & ... Cone cells, or cones, are one of three types of photoreceptor cells \textbf{in the retina} of mammalian eyes (e.g. the human eye). They are responsible for color vision and function best in  .. & \textbf{R:} in the retina 
       
       ~
       
       \textbf{R+C}: retina\\
       
       \hline
       
       
       
       who sang the theme song to step by step &
       ... \textbf{Jesse Frederick James Conaway} (born 1948), known professionally as Jesse Frederick, is an American film and television composer and singer best known for writing  ...
       & \textbf{R:} Jesse Frederick James Conaway
       
       \textbf{R+C:} Jesse Frederick James Conaway (born 1948), known professionally as Jesse Frederick\\
       
       \hline
    \end{tabular}
    \caption{Examples from the Natural Questions dev set wherein the answer corrector introduces an error in a previously correct reader output. The ground truth answer is marked in bold in each passage. \textbf{R} and \textbf{C} refer to reader and corrector, respectively.}
    \label{tab:example_predictions_wrong}
\end{table*}

Table~\ref{tab:predictions} shows the percentage of errors corrected in each error class. Corrections were made in all three categories, but more in 
\mbox{\emph{GT $\subset$ Prediction}}
and 
\mbox{\emph{Prediction $\cap$ GT $\neq \emptyset$}} 
than in 
\mbox{\emph{Prediction $\subset$ GT}}, indicating that the corrector learns the concepts of minimality and syntactic structure better than that of adequacy. We  note that most existing MRC systems that only output a single contiguous span are not equipped to handle multi-span discontinuous GT.

\section{Conclusion}
We describe a novel method for answer span correction in machine reading comprehension.
The proposed method operates by 
marking an original, possibly incorrect, answer prediction in context and then making a new prediction using a corrector model.
We show that this method corrects the predictions of a state-of-the-art English-language reader in different error categories.
In our experiments, the approach also generalizes well to multilingual and cross-lingual MRC in seven languages.
Future work will explore joint answer span correction and validation of the answerability of the question, re-using the original reader's output representations in the correction model and architectural changes enabling parameter sharing between the reader and the corrector.

\begin{table}[]
    \centering
    \begin{tabular}{|c | c  c|}
    \hline
    Error class&Total & Corrected \\
    \hline
    GT $\subset$ Prediction & 165 (28\%) &  62 (38\%) \\

    Prediction $\subset$ GT & 191 (33\%) & 18 (9\%)\\

    Prediction $\cap$ GT $\neq \emptyset$ & 37 (6\%) & 8 (22\%)\\
 
    Multi-span GT & 194 (34\%) & - \\
    \hline
    \end{tabular}
    
    \caption{Correction statistics for different error categories in 587 partial match (EM=0, F1$>$0) reader predictions.}
    \label{tab:predictions}
\end{table}



\section*{Acknowledgments}

We thank Tom Kwiatkowski for his help with debugging while submitting to the Google Natural Questions leaderboard. We would also like to thank the multilingual NLP team at IBM Research AI and the anonymous reviewers for their helpful suggestions and feedback.

\bibliographystyle{acl_natbib}
\bibliography{emnlp2020}

\end{document}